\documentclass{article}

\usepackage[final, nonatbib]{nips_2016} % CP - needed to compile
\usepackage[normalem]{ulem} % CP: added for strikethroughs

\usepackage{mathtools}
\usepackage{multicol}

\DeclarePairedDelimiterX{\infdivx}[2]{\Big(}{\Big)}{%
  #1\;\delimsize\|\;#2%
}
\newcommand{\infdiv}{D_{KL}\infdivx}

\usepackage[utf8]{inputenc} % allow utf-8 input
\usepackage[T1]{fontenc}    % use 8-bit T1 fonts
\usepackage{hyperref}       % hyperlinks
\usepackage{url}            % simple URL typesetting
\usepackage{booktabs}       % professional-quality tables
\usepackage{amsfonts}       % blackboard math symbols
\usepackage{nicefrac}       % compact symbols for 1/2, etc.
\usepackage{microtype}      % microtypography

\usepackage{amsmath}
\usepackage{xcolor}
\usepackage{graphicx}

\newcommand{\WW}{\mathbf{W}}

\newcommand{\BB}{\mathbf{B}}

\newcommand{\tW}{\textnormal{W}}
\newcommand{\RNN}{\textnormal{RNN}}

\newcommand{\hh}{\mathbf{h}}

\newcommand{\ee}{\mathbf{e}}
\newcommand{\cc}{\mathbf{c}}
\newcommand{\xx}{\mathbf{x}}
\newcommand{\rr}{\mathbf{r}}
\newcommand{\ff}{\mathbf{f}}
\newcommand{\ggg}{\mathbf{g}}
\newcommand{\uu}{\mathbf{u}}
\newcommand{\vv}{\mathbf{v}}
\newcommand{\aaa}{\mathbf{a}}
\newcommand{\zz}{\mathbf{z}}
\newcommand{\EE}{\mathbf{E}}

\newcommand{\yy}{\mathbf{y}}
\newcommand{\qq}{\mathbf{q}}

\newcommand{\gen}{\mbox{\scriptsize gen}}
\newcommand{\rate}{\mbox{\scriptsize rate}}
\newcommand{\fac}{\mbox{\scriptsize fac}}
\newcommand{\encf}{\mbox{\scriptsize enc, f}}
\newcommand{\encb}{\mbox{\scriptsize enc, b}}
\newcommand{\con}{\mbox{\scriptsize con}}

\title{LFADS - Latent Factor Analysis via Dynamical Systems}

\author{
  David Sussillo\thanks{also Department of Electrical Engineering, Stanford University} \\
  Google, Inc. \\
  \texttt{sussillo@google.com} \\
  \And
  Rafal Jozefowicz\thanks{Current address: OpenAI} \\
  Google, Inc. \\
  \texttt{rafal@openai.com} \\
  \AND
  L.F. Abbott \\
  Department of Neuroscience \\
  Columbia University \\
  \texttt{lfabbott@columbia.edu} \\
  \And
  Chethan Pandarinath\thanks{Current address: Coulter Department of Biomedical Engineering, Georgia Tech and Emory University}\\
  Department of Neurosurgery \\ 
  Stanford University \\
  \texttt{chethan@gatech.edu} \\
}

\begin{document}
\maketitle
%\begin{multicols}{2}

\begin{abstract}
  Neuroscience is experiencing a data revolution in which many hundreds or
  thousands of neurons are recorded simultaneously. Currently, there is little
  consensus on how such data should be analyzed.  Here we introduce LFADS
  (Latent Factor Analysis via Dynamical Systems), a method to infer latent
  dynamics from simultaneously recorded, single-trial, high-dimensional neural
  spiking data.  LFADS is a sequential model based on a variational
  auto-encoder. By making a dynamical systems hypothesis regarding the
  generation of the observed data, LFADS reduces observed spiking to a set of
  low-dimensional temporal factors, per-trial initial conditions, and inferred
  inputs.  We compare LFADS to existing methods on synthetic data and show that
  it significantly out-performs them in inferring neural firing rates and latent
  dynamics.
\end{abstract}

\section{Introduction}

Over the past two decades, the ability to record large numbers of neurons
simultaneously has increased dramatically, primarily through the use of
multi-unit electrode arrays and imaging technologies.  The change from
single-unit recordings to simultaneously recorded, high-dimensional data is
exciting, but analyzing this data presents significant challenges. The
traditional approach, necessitated by limited experimental observations, is to
trial-average recorded spike trains from single neurons, perhaps with some
temporal filtering, and then to construct peri-stimulus time histograms (PSTHs).
Despite many attempts to make sense of high-dimensional neuroscience data,
e.g. \cite{churchland2012neural,kato2015global,kobak2016demixed,mante2013context,rigotti2013importance},
the vast majority of experimental neuroscientists still use single-unit PSTHs in
their analyses.

The most obvious challenge to advancing neural data analysis is that brains are
immensely complex, and they solve multiple and varied problems ranging from
vision to motor planning and navigation.  Given the generality of the problems
brains must solve, it is not surprising that no one-size-fits-all algorithm has
yet materialized. Two ends of a spectrum for generic computation are
feed-forward processing and sequential processing.  In the feed-forward case,
temporal dynamics play no role, and computation is purely input driven.  In the
sequential case, on the other hand, dynamics are paramount, and computation
results from an autonomous nonlinear dynamical system starting from a specific
initial condition and run for some period of time to produce ``computational
dynamics''.

Here we develop a deep learning architecture, called \textbf{Latent Factor
  Analysis via Dynamical Systems} (\textbf{LFADS}; {\it ``ell-fads''}), which
lies further towards the sequential end of this spectrum, while still allowing
for external inputs and input driven dynamics.  LFADS implements the hypothesis
that a driven nonlinear dynamical system provides a reasonable model of many
neural processes.  The primary goal of LFADS is to infer smooth dynamics from
recorded neural spike trains on a single-trial basis.  In doing so, LFADS
provides a number of other components that can aid in understanding the data
better: a low-dimensional set of temporal factors that explain the observed
spike trains, a recurrent network that produces the smoothed data and can be
analyzed using techniques such as those found in \cite{sussillo2013opening}, a
set of initial conditions that can be used as a code for each trial (if the data
is broken up into trials) and, finally, LFADS infers inputs.  Inferring inputs
is predicated on the idea that a dynamical system provides a notion of surprise,
namely, if a powerful nonlinear dynamical system cannot generate the data then,
necessarily, an external perturbation to the system must have occurred.  Within
the LFADS architecture, we learn this perturbation and identify it as an
inferred input.

The ability of LFADS to infer input is an extremely useful feature for
neuroscience research.  Even extensive neural recordings cover only a tiny
fraction of an animal's neurons, which means that the recorded neurons receive
input from other neurons that are not recorded.  Unmeasured input introduces a
major source of uncertainty into our understanding and modeling of neural
circuits.  Even in cases where we know that input is affecting a particular
circuit, for example inputs due to sensory stimuli presented to the animal
during the recording session, we typically do not know the form that this input
takes.  LFADS is designed to infer such inputs on the basis of the recorded data
alone.

\section{The LFADS Model}

The LFADS model is an instantiation of a variational auto-encoder (VAE)
\cite{kingma2013auto, rezende2014stochastic} extended to sequences, as in
\cite{gregor2015draw} or \cite{krishnan2015deep}.  The VAE consists of two
components, a decoder or generator and an encoder.  The generator assumes that
data, denoted by $\xx$, arise from a random process that depends on a vector of
stochastic latent variables $\zz$, samples of which are drawn from a prior
distribution $P(\zz)$. Simulated data points are then drawn from a conditional
probability distribution, $P(\xx|\zz)$ (we have suppressed notation reflecting
the dependence on parameters of this and the other distributions we discuss).

The VAE encoder transforms actual data vectors, $\xx$, into a conditional
distribution over $\zz$, $Q(\zz|\xx)$. $Q(\zz|\xx)$ is a trainable approximation
of the posterior distribution of the generator, $Q(\zz|\xx) \approx P(\zz|\xx) =
P(\xx|\zz)P(\zz)/P(\xx)$. $Q(\zz|\xx)$ can also be thought of as an encoder from
the data to a data-specific latent code $\zz$, which can be decoded using the
generator (decoder).  Hence the autoencoder; the encoder $Q$ maps the actual
data to a latent stochastic ``code", and the decoder $P$ maps the
latent code back to an approximation of the data.  Specifically, when the two
parts of the VAE are combined, a particular data point is selected and an
associated latent code, $\hat{\zz}$ (we use $\hat{\zz}$ to denote a sample of
the stochastic variable $\zz$) is drawn from $Q(\zz|\xx)$. A data sample is then
drawn from $P(\xx|\hat{\zz})$, on the basis of the sampled latent variable.  If
the VAE has been constructed properly, $\hat{\xx}$ should resemble the original
data point $\xx$.

The loss function that is minimized to construct the VAE involves
minimizing the Kullbach-Liebler divergence between the encoding distribution
$Q(\zz|\xx)$ and the prior distribution of the generator, $P(\zz)$, over all
data points.  Thus, if training is successful, these two distributions should
converge and, in the end, statistically accurate simulations of the data can be
generated by running the generator model alone.

We now translate this general description of the VAE into the specific LFADS
implementation.  Borrowing notation from \cite{gregor2015draw}, we denote an
affine transformation from a variable $\uu$ to a variable $\vv$ as $\vv =
\tW(\uu)$, we use $[\cdot, \cdot]$ to represent vector concatenation, and we
denote a temporal update of a recurrent neural network receiving an input as
$\mbox{state}_{t} = \textnormal{RNN}(\mbox{state}_{t-1}, \mbox{input}_t)$.

\subsection{LFADS Generator}

The neural data we consider, $\xx_{1:T}$, consists of spike trains from $D$
recorded neurons.  Each instance of a vector $\xx_{1:T}$ is referred to as a
trial, and trials may be grouped by experimental conditions, such as stimulus or
response types.  The data may also include an additional set of observed
variables, $\aaa_{1:T}$, that may refer to stimuli being presented or other
experimental features of relevance.  Unlike $\xx_{1:T}$, the data described by
$\aaa_{1:T}$ is not itself being modeled, but it may provide important
information relevant to the modeling of $\xx_{1:T}$.  This introduces a slight
complication: we must distinguish between the complete data set, $\{\xx_{1:T},
\aaa_{1:T}\}$ and the part of the data set being modeled, $\xx_{1:T}$.  The
conditional distribution of the generator, $P(\xx|\zz)$, is only over $\xx$,
whereas the approximate posterior distribution, $Q(\zz|\xx, \aaa)$, depends on
both types of data.

LFADS assumes that the observed spikes described by $\xx_{1:T}$ are samples from
a Poisson process with underlying rates $\rr_{1:T}$.  Based on the dynamical
systems hypothesis outlined in the Introduction, the goal of LFADS is to infer a
reduced set of latent dynamic variables, $\ff_{1:T}$, from which the firing
rates can be constructed.  The rates are determined from the factors by an
affine transformation followed by an exponential nonlinearity, $\rr_{1:T} =
\exp(W^{\rate}(\ff_{1:T}))$.  The choice of a low-d representation for the
factors is based on the observation that the intrinsic dimensionality of neural
recordings tends to be far lower than the number of neurons recorded, e.g.
\cite{churchland2012neural,kato2015global,mante2013context}, and see \cite{gao2015simplicity}
for a more complete discussion.

The factors are generated by a recurrent nonlinear neural network and are
characterized by an affine transformation of its state vector, $\ff_{1:T} =
W^{\fac}(\ggg_{1:T})$.  Running the network requires an initial condition
$\ggg_0$, which is drawn from a prior distribution $P(\ggg_0)$.  Thus, $\ggg_0$
is part of the stochastic latent variable $\zz$ discussed above.

There are different options for sources of input to the recurrent generator
network, First, as in some of the examples to follow, the network may receive no
input at all.  Second, it may receive the information contained in the
non-modeled part of the data, $\aaa_{1:T}$, in the form of a network input.  We
discuss this option in the Discussion, but we do not implement it here.  Instead,
as a third option, we introduce an inferred input $\uu_{1:T}$.  When an inferred
input is included, the stochastic latent variable is expanded to include it,
$\zz = \{\ggg_0, \uu_{1:T}\}$.  At each time step, $\uu_t$ is drawn from a prior
distribution $P(\uu_t)$.

\begin{figure*}
\centering
\includegraphics[width=14cm]{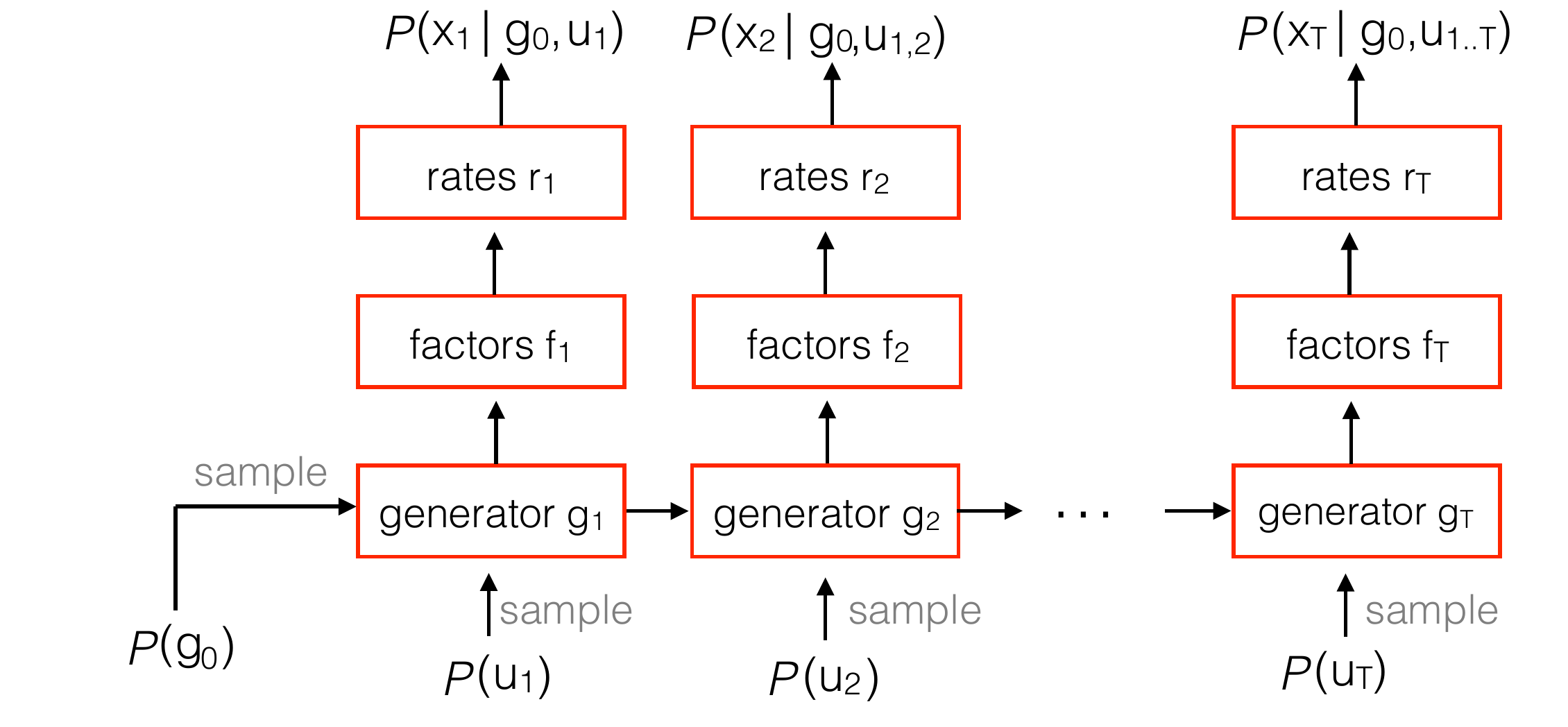}
\caption{ \textbf{The LFADS generator.} The generative LFADS model is a
  recurrent network with a feed-forward readout. The generator takes a sampled initial condition, $\hat{\ggg}_0$ and a sampled input, $\hat{\uu}_t$, at each time step, and iterates forward.  At each time step the temporal factors, $\ff_t$, and the rates, $\rr_t$ are
  generated in a feed-forward manner from $\ggg_t$.  Spikes are generated from a Poisson process, $\hat{\xx}_t \sim\ \textnormal{Poisson}(\xx_t|\rr_t)$.  The initial condition
  and inputs are sampled from diagonal Gaussian distributions with zero mean and
  fixed chosen variance. }
\label{fig_lfads_generative_model}
\end{figure*}

The LFADS generator with inferred input is thus described by the following
procedure and equations.  First an initial condition for the generator is
sampled from the prior
\begin{align}
  \hat{\ggg}_0 &\sim\ P(\ggg_0). \label{g0_prior_sample}
\end{align}
At each time step $t=1,\ldots, T$, an inferred input, $\hat{\uu}_t$, is sampled
from its prior and fed into the network, and the network is evolved forward in
time,
\begin{align}
  \hat{\uu}_t &\sim\ P\left(\uu_t\right) \label{u_t_prior_sample}\\
  \ggg_t &= \RNN^{\gen}\left(\ggg_{t-1}, \hat{\uu}_t\right) \label{recurrGen}\\
  \ff_t &= \WW^{\fac}(\ggg_t) \\
  \rr_t &= \exp\left(\tW^{\rate}\left(\ff_t\right)\right) \\
  \hat{\xx}_t &\sim\ \textnormal{Poisson}(\xx_t|\rr_t).
\end{align}
Here ``Poisson" indicates that each component of the spike vector $\xx_t$ is
generated by an independent Poisson process at a rate given by the corresponding
component of the rate vector $\rr_t$.  The priors for both $\ggg_0$ and $\uu_t$,
$P(\cdot)$, are diagonal Gaussian distributions with zero mean and a fixed
chosen variance (see Appendix).  We chose the GRU \cite{chung2014empirical} as
our recurrent function for all the networks we use (defined in the Appendix).
We have not included the observed data $\aaa$ in the generator model defined
above, but this can be done simply by including $\aaa_t$ as an additional input
to the recurrent network in equation~\ref{recurrGen}.  Note that doing so will
make the generation process necessarily dependent on including an observed input.
The generator model is illustrated in Fig.~\ref{fig_lfads_generative_model}.
This diagram and the above equations implement the conditional distribution
$P(\xx|\zz) = P(\xx|\{\ggg_0, \uu_{1:T}\})$ of the VAE decoder.

\begin{figure*}
\centering
\includegraphics[width=14cm]{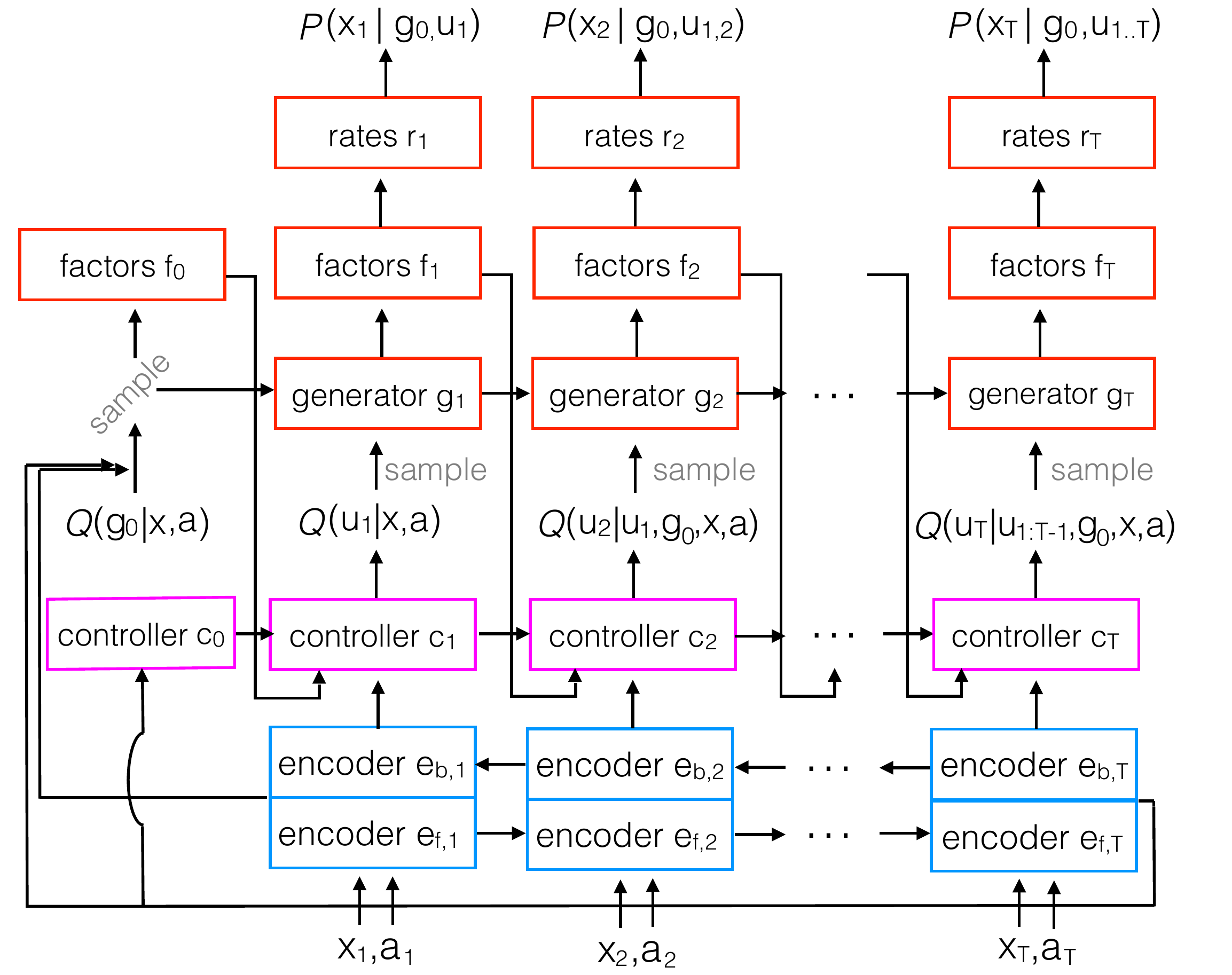}
\caption{ \textbf{The full LFADS model for inference.} The decoder portion is in
  red, the encoder portion in blue and the controller in purple.  To infer the
  latent dynamics from the recorded neural spike trains $\xx_{1:T}$ and other
  data $\aaa_{1:T}$, initial conditions for the controller and generator
  networks are encoded from inputs.  In the case of the generator, the initial
  condition $\hat{\ggg}_0$ is drawn from an approximate posterior
  $Q(\ggg_0|\xx_{1:T}, \aaa_{1:T})$ (in this figure, for compactness, we use
  $\xx$ and $\aaa$ to denote $\xx_{1:T}$ and $\aaa_{1:T}$).  The low-dimensional
  factors at $t=0$, $\ff_0$, are computed from $\hat{\ggg}_0$.  The controller
  initial condition, $\cc_0$, is also generated from $\EE$.  The controller then
  propagates one step forward in time, receiving the sample factors $\ff_0$ as
  well as bidirectionally encoded inputs $\tilde\EE_1$ computed from $\xx_{1:T},
  \aaa_{1:T}$ (again for compactness, we show a single encoder, though there is
  one for the initial conditions, and one for the inferred inputs).  The
  controller produces, through an approximate posterior $Q(\uu_1|\ggg_0,
  \xx_{1:T}, \aaa_{1:T})$, a sampled inferred input $\hat{\uu}_1$ that is fed
  into the generator network.  The generator network then produces $\{\ggg_1,
  \ff_1, \rr_1\}$, with $\ff_1$ the factors and $\rr_1$ the Poisson rates at $t
  = 1$. The process continues iteratively so, at time step $t$, the generator
  network receives $\ggg_{t-1}$ and $\hat{\uu}_t$ sampled from
  $Q(\uu_t|\uu_{1:t-1}, \ggg_0, \xx_{1:T}, \aaa_{1:T})$.  The job of the
  controller is to produce a nonzero inferred input only when the generator
  network is incapable of accounting for the data autonomously. Although the
  controller is technically part of the encoder, it is run in a forward manner
  along with the decoder.}
\label{fig_lfads_full_model}
\end{figure*}

\subsection{LFADS Encoder}

The approximate posterior distribution for LFADS is the product of two
conditional distributions, one for $\ggg_0$ and one for $\uu_t$.  Both of these
distributions are diagonal Gaussians with means and variances determined by the
outputs of the encoder or controller RNNs (see Fig.~\ref{fig_lfads_full_model}
and below).  We begin by describing the network that defines $Q(\ggg_0|\xx,
\aaa)$.  Its mean and variance are given in terms of a variable $\EE$ by
\begin{align}
  \mu^{g_0} &= \tW^{\mu^{g_0}}(\EE) \\
  \sigma^{g_0} &= \exp\left(\frac{1}{2}\tW^{\sigma^{g_0}}(\EE)\right).
\end{align}
$\EE$ is obtained by running a recurrent network both forward (from $t=1$ to
$t=T$) and backward (from $t=T$ to $t=1$) in time,
\begin{align}
  \ee^b_t &= \RNN^{\encb}\left(\ee^b_{t+1}, \left[\xx_t, \aaa_t\right]\right) \label{eqn_enc_1} \\ 
  \ee^f_t &= \RNN^{\encf}\left(\ee^f_{t-1}, \left[\xx_t, \aaa_t\right]\right) 
  \label{eqn_enc_2}
  \end{align}
with $\ee^b_{T+1}$ and $\ee^f_0$ learnable parameters.  Once this is done, $\EE$ is the concatenation 
\begin{align}
\EE&= \left[\ee^b_1, \ee^f_T\right] \label{eqn_enc_3}. 
 \end{align}
Running the encoding network both forward and backward in time allows $\EE$ to
reflect the entire time history of the data $\xx_{1:T}$ and $\aaa_{1:T}$.

The approximate posterior distribution for $\uu_t$ is defined in a more complex
way that involves both an encoder network and another RNN called the controller.
The encoder network with state variables $\tilde{\ee}^b$ and $\tilde{\ee}^f$ is
described by equations identical to~\ref{eqn_enc_1} and~\ref{eqn_enc_2}
(although with different trainable network parameters), and these serve to
define the time-dependent variable
\begin{align}
\tilde{\EE}_t = \left[\tilde{\ee}^b_t, \tilde{\ee}^f_t\right]  .
\end{align}
Rather than feeding directly into a Gaussian distribution, this variable is
passed through the controller RNN, which also receives the latent dynamic
factor, $\ff_{t-1}$ as input,
\begin{align}
  \cc_t &= \RNN^{\con}\left(\cc_{t-1}, \left[\tilde{\EE}_{t}, \ff_{t-1}\right]\right) .
\end{align}
Thus, the controller is privy to the information about $\xx_{1:T}$ and
$\aaa_{1:T}$ encoded in the variable $\tilde\EE_t$, and it receives information
about what the generator network is doing through the latent dynamic factor
$\ff_{t-1}$.  It is necessary for the controller to receive the factors so that
it can correctly decide when to intervene in the generation process.  Because
$\ff_{t-1}$ depends on both $\ggg_0$ and $\uu_{1:t-1}$, these stochastic
variables are included in the conditional dependence of the approximate
posterior distribution $Q(\ggg_0, \uu_t| \uu_{1:t-1}, \ggg_0, \xx_{1:T},
\aaa_{1:T})$.  The initial state of the controller network, $\cc_0$, is defined
deterministically, rather than drawn from a distribution, by the affine
transformation
\begin{align}
c_0 &= \tW^{c_0}(\EE) .
\label{cZero}
\end{align}
$\EE$ in this expression is the same variable that determines the mean and the variance of the distribution from which $\ggg_0$ is drawn, as defined by equations~\ref{eqn_enc_1}-\ref{eqn_enc_3}.

Finally, the inferred input, $\uu_t$, at each time, is a stochastic variable drawn from 
a diagonal Gaussian distribution with mean and log-variance given by an
affine transformation of the controller network state,  $\cc_t$,
\begin{align}
 \hat{\uu}_t &\sim\ Q\left(\uu_t\;|\;\mu^u_t, \sigma^u_t\right) \label{u_t_sample}
 \end{align}
with
\begin{align}
  \mu^u_t &= \tW^{\mu^{u}}(\cc_t) \\
  \sigma^u_t &= \exp\left(\frac{1}{2}\tW^{\sigma^u}(\cc_t)\right).
\end{align}
We control the information flow out of the controller and into the generator by
applying a regularizer on $\uu_t$ (a KL divergence term, described below and also
in the Appendix), and also by explicitly limiting the dimensionality of $\uu_t$,
the latter of which is controlled by a hyper parameter.

\subsection{The full LFADS inference model}

The full LFADS model (Fig.~\ref{fig_lfads_full_model}) is run in the following
way.  First, a data trial is chosen, the encoders are run, and an initial
condition is sampled from the approximate posterior $Q(\ggg_0|\xx_{1:T},
\aaa_{1:T})$.  An initial condition for the controller is also sampled
(equation~\ref{cZero}).  Then, for each time step from 1 to $T$, the generator
is updated, as well as the factors and rates, according to
\begin{align}
  \cc_t &= \RNN^{\con}\left(\cc_{t-1}, \left[\tilde{\EE}_{t}, \ff_{t-1}\right]\right) \label{eqn_dropout_2}\\
  \mu^u_t &= \tW^{\mu^{u}}(\cc_t) \\
  \sigma^u_t &= \exp\left(\frac{1}{2}\tW^{\sigma^u}(\cc_t)\right) \\
  \hat{\uu}_t &\sim\ Q\left(\uu_t\;|\;\mu^u_t, \sigma^u_t\right) \\
  \ggg_t &= \RNN^{\gen}\left(\ggg_{t-1}, \hat{\uu}_t\right) \label{eqn_dropout_3} \\
  \ff_t &= \WW^{\fac}(\ggg_t)  \\
  \rr_t &= \exp\left(\tW^{\rate}\left(\ff_t\right)\right) \\
  \hat{\xx}_t &\sim\ \textnormal{Poisson}(\xx_t|\rr_t).
\end{align}

After training, the full model can be run, starting with any single trial or a
set of trials corresponding to a particular experimental condition to determine
the associated dynamic factors, firing rates and inferred inputs for that trial
or condition.  This is done by averaging over several runs to marginalize over
the stochastic variables $\ggg_0$ and $\uu_{1:T}$.
% and, if necessary, over the stochastic spiking output of the generator.  The generator can also be run independently to produce simulated data.
\subsection{The loss function}

To optimize our model, we would like to maximize the log likelihood of the data,
$\sum_{\xx} \log P(\xx_{1:T})$, marginalizing over all latent variables.  For
reasons of intractability, the VAE framework is based on maximizing a lower
bound, $\mathcal{L}$, on the marginal data log-likelihood,
\begin{align}
  \log P(\xx_{1:T}) \geq \mathcal{L} = \mathcal{L}^x - \mathcal{L}^{KL} .
\end{align}
$\mathcal{L}^x$ is the log-likelihood of the reconstruction of the data, given the inferred firing rates, and $\mathcal{L}^{KL}$ is a non-negative penalty that restricts the approximate posterior distribution for deviating too far from the prior distribution.  These are defined as
\begin{align}
  \mathcal{L}^x &= \left\langle\sum_{t=1}^T \log\Big( \textnormal{Poisson}(\xx_t | \rr_t)\Big)\right\rangle_{\ggg_0, \uu_{1:T}} \\ 
  \mathcal{L}^{KL} &= \left\langle\infdiv{ Q\left( \ggg_0|\xx_{1:T}, \aaa_{1:T} \right)} {P\left(\ggg_0\right)}\right\rangle_{\!\ggg_0, \uu_{1:T}} \;\;\;+ \nonumber \\ 
  &\phantom{=} \left\langle\sum_{t=1}^T \infdiv{Q\left(\uu_t|\uu_{1:t-1}, \ggg_0, \xx_{1:T}, \aaa_{1:T}\right)}{P\left(\uu_t\right)}\right\rangle_{\!\!\ggg_0, \uu_{1:T}}, \label{eq_L_KL}
\end{align}
where the brackets denote marginalizations over the subscripted variables. Evaluating the $T+1$ KL terms can be done analytically for the Gaussian distributions we use; the formulae are found in Appendix B of \cite{kingma2013auto}.  We minimize the negative bound, $-\mathcal{L}$, using the reparameterization trick to back-propagate low-variance, unbiased gradient estimates.  These gradients are used to train the system in an end-to-end fashion, as is typically done in deterministic settings.

\section{Relation to Previous Work}

Recurrent neural networks have been used extensively to model neuroscientific
data (e.g. \cite{sussillo2009generating} \cite{mante2013context},
\cite{carnevale2015dynamic}, \cite{sussillo2015neural},
\cite{rajan2016recurrent}), but the networks in these studies were all trained
in a deterministic setting.  An important recent development in deep learning
has been the advent of the variational auto-encoder \cite{kingma2013auto} \cite{
  rezende2014stochastic}, which combines a probabilistic framework with the
power and ease of optimization of deep learning methods.  VAEs have since been
generalized to the recurrent setting, for example with variational recurrent
networks \cite{chung2015recurrent}, deep Kalman filters \cite{krishnan2015deep},
and the RNN DRAW network \cite{gregor2015draw}.

There is also a line of research applying probabilistic sequential graphical
models to neural data.  Recent examples include PLDS \cite{macke2011empirical},
switching LDS \cite{petreska2011dynamical}, GCLDS \cite{gao2015high}, and PfLDS
\cite{gao2016linear}.  These models employ a linear Gaussian dynamical system
state model with a generalized linear model (GLM) for the emissions
distribution, typically using a Poisson process.  In the case of the switching
LDS, the generator includes a discrete variable that allows the model to switch
between linear dynamics.  GCLDS employs a generalized count distribution for the
emissions distribution. Finally, in the case of PfLDS, a nonlinear feed-forward
function (neural network) is inserted between the LDS and the GLM.
% \TODO{Is David Pfau's work relevant?} \cite{pfau2013robust}

Gaussian process models have also been explored.  GPFA \cite{yu2009gaussian}
uses Gaussian processes (GPs) to infer a time constant with which to smooth
neural data and has seen widespread use in experimental laboratories.  More
recently, the authors of \cite{zhao2016variational} have used a variational
approach (vLGP) to learn a GP that then passes through a nonlinear feed-forward
function to extract the single-trial dynamics underlying neural spiking data.

The authors of \cite{krishnan2015deep} have defined a very general nonlinear
variational sequential model, which they call the Deep Kalman Filter (DKF).  The
authors then apply a DKF to a synthetic ``healing MNIST'' task.  Due to the
generality of the DKF equations, LFADS is likely one of many possible
instantiations of a DKF applied to neural data (in the same sense that a
convolutional network architecture applied to images is also a feed-forward
network, for example).
%% However, signicant
%% differences remain, for example, the flow of information through the controller
%% in LFADS. If $\ggg_0$ is undefined, and we fed only the observed inputs
%% $\aaa_t$, and not $\xx_t$, into LFADS and made the dimension of $\uu_t$ the same
%% as $\xx_t$, then LFADS would be similar to DKF\@.

Like a Kalman filter, LFADS decomposes data into a set of inferred dynamics and
a set of innovation-like quantities that help explain the observed data when the
state model cannot.  Recasting our work in the language of Kalman filters, our
nonlinear generator is analogous to the linear state estimator in a Kalman
filter, and we can think of the inferred inputs in LFADS as innovations in the
Kalman filter language.  However, an ``LFADS innovation'' is not strictly
defined as an error between the measurement and the readout of the state
estimate.  Rather, the LFADS innovation may depend on the observed data and the
generation process in extremely complex ways.  

%% Finally, unlike the Kalman
%% filter, the architecture of LFADS is explicitly constructed to decompose the
%% observed dynamics by funneling information into the high-dimensional
%% initial condition, $\ggg_0$, thus forcing the generator to autonomously produce
%% state trajectories consistent with the data. As a backup, data variance can be
%% explained by the low-dimensional stochastic inferred input.

%% The departure here
%% from the Kalman filter is important.  LFADS provides a full state specification
%% while also allowing for a less informative input (both low-dimensional and
%% subject to a KL divergence penalty with an uninformative prior).  While $\ggg_0$
%% is also subject to a KL divergence penalty, the fact that $\ggg_0$ is
%% high-dimensional makes this a significantly less severe restriction than that on
%% the inferred input.

\section{Results}

We compared the performance of LFADS to three existing methods that estimate
latent state from neural data: GPFA \cite{yu2009gaussian}, vLGP
\cite{zhao2016variational}, and PfLDS \cite{gao2016linear}.  To test LFADS and
to compare its performance with other approaches, we generated synthetic
stochastic spike trains from two different deterministic nonlinear systems. The
first is the standard Lorenz system.  The second is a recurrent nonlinear neural
network that we call the data RNN (to distinguish it from the RNNs within
LFADS).  For these two systems, we compared LFADS without any inferred input
(because none is needed) with other methods.  We also added an input to the data
RNN to highlight the inferred input feature of LFADS.  As shown below, in the
first two cases, LFADS outperforms the other methods, and in the third case,
LFADS is able to infer the timing of the actual input quite reliably.

\subsection{Lorenz system}

The Lorenz system is a set of nonlinear equations for three dynamic
variables. Its limited dimensionality allows its entire state space to be
visualized. The evolution of the system's state is governed as follows
\begin{align}
  \dot{y}_1 &= \sigma\left(y_2 - y_1\right) \\
  \dot{y}_2 &= y_1 (\rho - y_3) - y_2 \\ 
  \dot{y}_3 &= y_1 y_2 - \beta y_3 .
\end{align}
We used standard parameter values $\sigma=10$, $\rho=28$, and $\beta=8/3$, and
used Euler integration with $\Delta t = 0.006$.  As in
\cite{zhao2016variational}, we simulated a population of neurons with firing
rates given by linear readouts of the Lorenz variables using random weights,
followed by an exponential nonlinearity. Spikes from these firing rates were
then generated by a Poisson process. We then tested the ability of each method
to infer the latent dynamics of the Lorenz system (i.e., the values of the three
dynamic variables) from the spiking activity alone.

\begin{figure*}
\centering
\includegraphics[width=14cm,trim={0 1.5cm 0 1.5cm},clip]{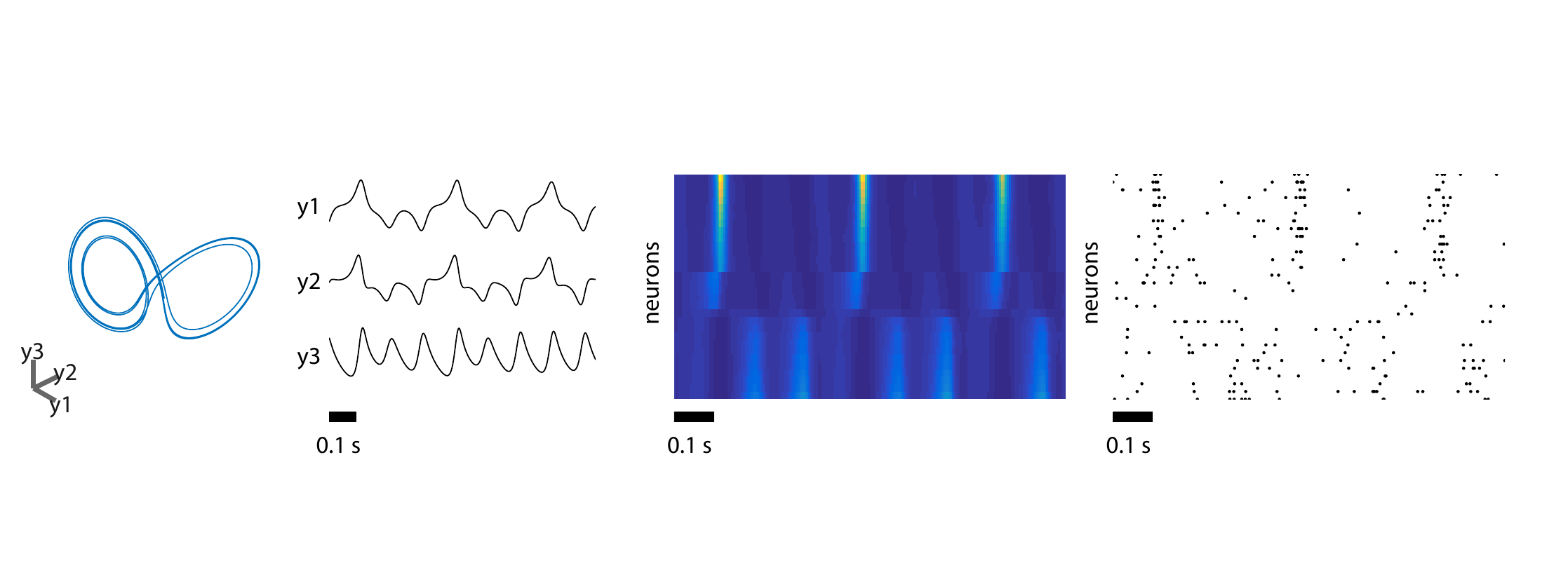}
\caption{ \textbf{Overview of Lorenz task.}  An example trial illustrating the
  evolution of the Lorenz system in its 3-dimensional state space (far left) and its dynamic variables as  a function of time (left middle).  Firing rates for the 30 simulated neurons are computed by a linear readout of the latent variables followed by an exponential nonlinearity (right middle; with neurons sorted according to their weighting for the first Lorenz dimension). Spike times for the neurons are shown at the right. }
\label{fig_lorenz_overview}
\end{figure*}

Our synthetic dataset consisted of 65 conditions, with 20 trials per condition.
Each condition was obtained by starting the Lorenz system with a random initial
state vector and running it for 1s. Twenty different spike trains were then
generated from the firing rates for each condition.  Models were trained using
80\% of the data (16 trials/condition) and evaluated using 20\% of the data (4
trials/condition). Fig.~\ref{fig_lorenz_overview} shows the Lorenz variables and
spike times for an example trial. While this simulation is structurally quite
similar to the Lorenz system used in \cite{zhao2016variational}, we purposefully
chose parameters that made the dataset more challenging. Specifically, relative
to \cite{zhao2016variational}, we limited the number of observations to 30
simulated neurons instead of 50, decreased the baseline firing rate from 15
spikes/sec to 5 spikes/sec, and sped up the dynamics by a factor of 4.

\begin{figure*}
\centering
\includegraphics[width=14cm,trim={1.5cm 0 1.5cm 0},clip]{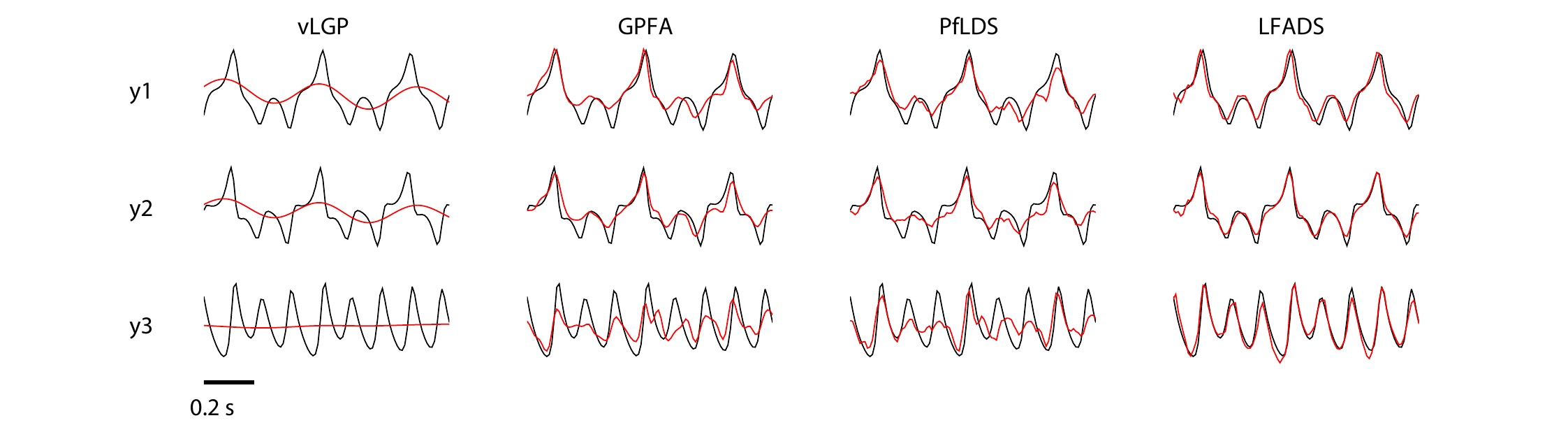}
\caption{ \textbf{Sample performance for each method applied to Lorenz-based spike trains.} Each panel shows actual (black traces) and inferred (red trace) values of
  the three latent variables for a single example trial.}
\label{fig_lorenz_rates}
\end{figure*}
\begin{table}[t]
\centering % used for centering table
\begin{tabular}{c c c c c} % centered columns (5 columns)
\hline\hline \\ [1ex] %inserts double horizontal lines 
Dim & vLGP & GPFA & PfLDS & LFADS \\ [0.5ex] % inserts table
%heading
\hline % inserts single horizontal line
y1 & 0.477 & 0.713 & 0.784 & \textbf{0.826} \\ % inserting body of the table
y2 & 0.377 & 0.725 & 0.732 & \textbf{0.900} \\
y3 & 0.015 & 0.325 & 0.368 & \textbf{0.821} \\  [1ex] % [1ex] adds vertical space
\hline %inserts single line
\end{tabular}
\caption{$R^2$ results for each method applied to Lorenz-based spike trains.
  Compared methods are Variational Latent Gaussian Process
  (vLGP\cite{zhao2016variational}), Gaussian Process Factor Analysis
  (GPFA\cite{yu2009gaussian}), and Poisson Feed-forward neural network Linear
  Dynamical System (PfLDS\cite{gao2016linear}).  LFADS recovers more variance of
  the latent Lorenz dynamics, as measured by $R^2$ between the linearly
  transformed output of each model, and the dynamics of the latent Lorenz
  dimensions.}
\label{table_lorenz} 
\end{table}

Fig.~\ref{fig_lorenz_rates} shows the Lorenz latent dynamics for an example
trial and the latent dynamics inferred by each method (for LFADS, posterior
means averaged over 128 samples of $\ggg_0$ conditioned on the particular input
sequence). For each method, the inferred latents linearly transformed to the
actual latents to facilitate direct comparison. As shown, LFADS accurately
recovered the latent dynamics underlying the observed spike trains, consistently
outperforming the three other methods. We quantify this using $R^2$, i.e., the
fraction of the variance of the actual latent variables captured by the
estimated latent values (Table \ref{table_lorenz}).

\subsection{Inferring the dynamics of a chaotic RNN}
Next we tested the performance of each method at inferring the dynamics
of a more complex nonlinear dynamical system, a fully recurrent neural
network with strong coupling between the units.  We generated a synthetic
dataset from an $N$-dimensional continuous time nonlinear, so-called, ``vanilla"
RNN,
\begin{align}
  \tau\dot{\yy}(t) &= -\yy(t) + \gamma \WW^y\tanh(\yy(t)) + \BB\qq(t) .
\end{align}
This makes a compelling synthetic case study for our method because many recent
studies of neuroscientific data have used vanilla RNNs as their modeling tool
(e.g. \cite{sussillo2009generating} \cite{mante2013context},
\cite{carnevale2015dynamic}, \cite{sussillo2015neural},
\cite{rajan2016recurrent}).  It should be stressed that the vanilla RNN used as
the data RNN here does not have the same functional form as the network
generator used in the LFADS framework, which is a GRU (see Appendix). In this
example, we set $\BB = \qq = 0$, but we include an input in the following
section.

\begin{figure*}
\centering
\includegraphics[width=14cm,trim={0 0.5cm 3.5cm 1.5cm},clip]{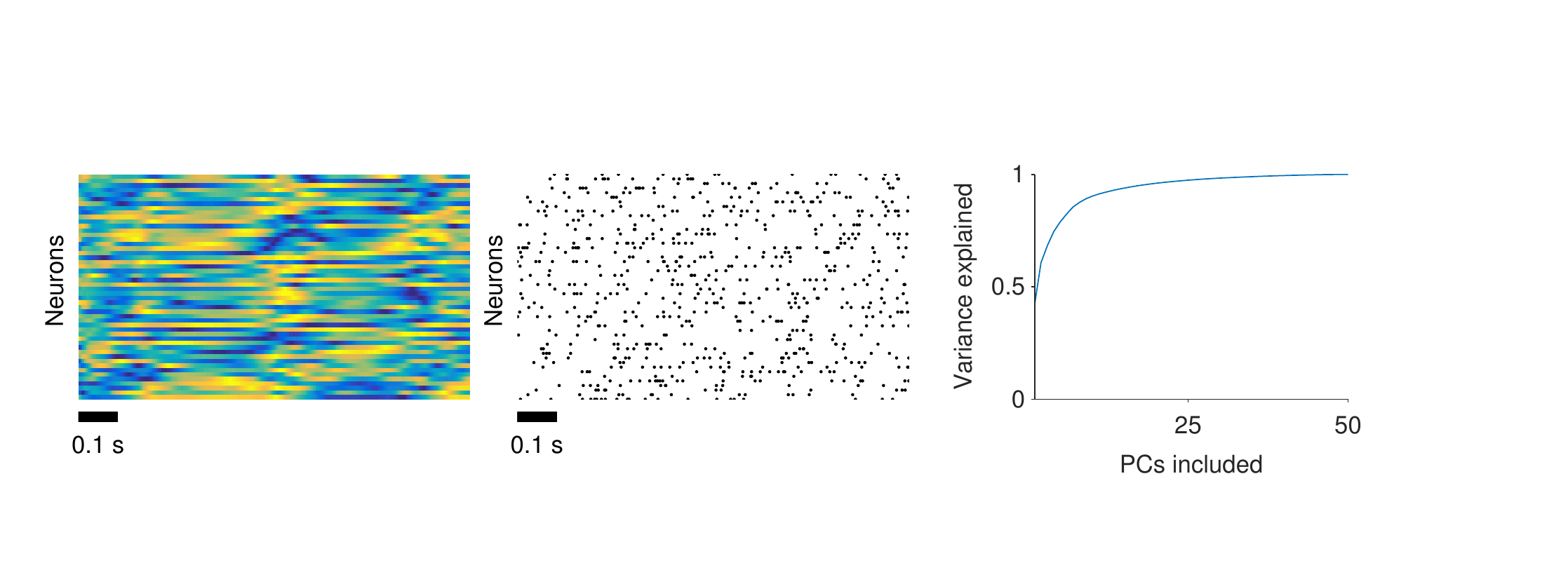}
\caption{ \textbf{Overview of RNN task.}  Firing rates generated on one example
  trial by the chaotic data RNN (left), and the resulting spike times for the
  simulated neurons (middle).  At the right, we show the fraction of explained
  variance of the simulated firing rates as a function of the number of
  principal components used. 20 PCs account for $> 95\%$ of the variance.}
\label{fig_rnn_no_thit_overview}
\end{figure*}

The elements of the matrix $\WW^y$ were drawn independently from a normal
distribution with zero mean and variance $1/N$ . We set $\gamma=2.5$, which
produces chaotic dynamics at a relatively slow timescale compared to $\tau$ (see
\cite{sussillo2009generating} for more details). Specifically, we set $N=50$,
$\tau=0.025$ s and used Euler integration with $\Delta t = 0.001$ s.  Spikes
were generated by a Poisson process with firing rates obtained by shifting and
scaling $\tanh(\yy(t))$ to give rates lying between 0 and 30 spikes/s (Fig.~
\ref{fig_rnn_no_thit_overview}).

Our dataset consisted of 400 conditions obtained by starting the data RNN at
different initial states with elements drawn from a normal distribution with
zero mean and unit variance.  Firing rates were then generated by running the
data RNN for 1 s, and 10 spiking trials were produced for each condition. Models
were trained using 80\% of the data (8 trials/condition) and evaluated using
20\% of the data (2 trials/condition).  Fig.~\ref{fig_rnn_no_thit_overview}
shows underlying firing rates and spiking activity for a single trial.

We used principal components analysis to determine the number of latent
variables needed to describe the data.  As expected, the state
of the data RNN has lower dimension than its number of neurons, and 20 principal
components are sufficient to capture $> 95\%$ of the variance of the system
(Fig.~\ref{fig_rnn_no_thit_overview}). We therefore restricted the latent space
to 20 dimensions for each of the models tested and, in the case of LFADS, set
the number of temporal factors $\ff_t$ to 20 as well.

\begin{figure*}
\centering
\includegraphics[width=14cm,trim={1.5cm 0 1.5cm 0},clip]{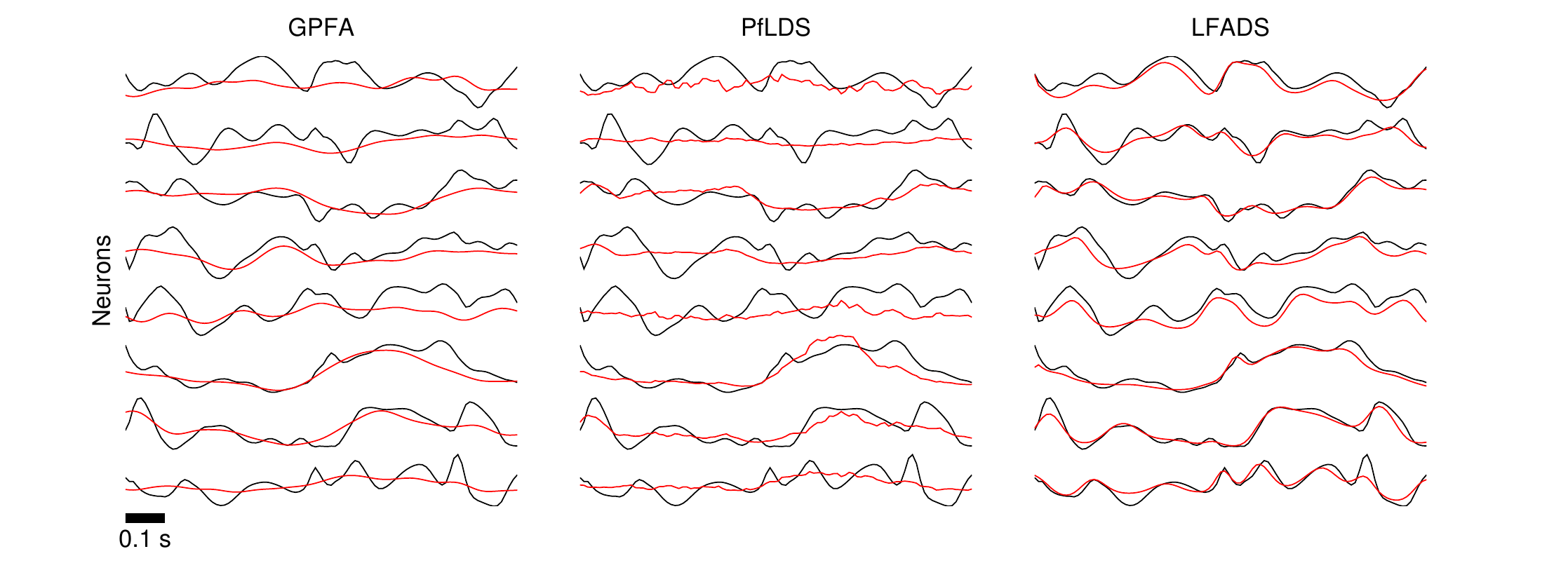}
\caption{ \textbf{Sample performance for each method on the RNN task.} Eight of
  the actual data RNN firing rates used to generate the observed spike trains
  are shown in black, and the corresponding inferred rates are in red. }
\label{fig_rnn_no_thit_wiggles}
\end{figure*}

\begin{figure*}
\centering
\includegraphics[width=14cm]{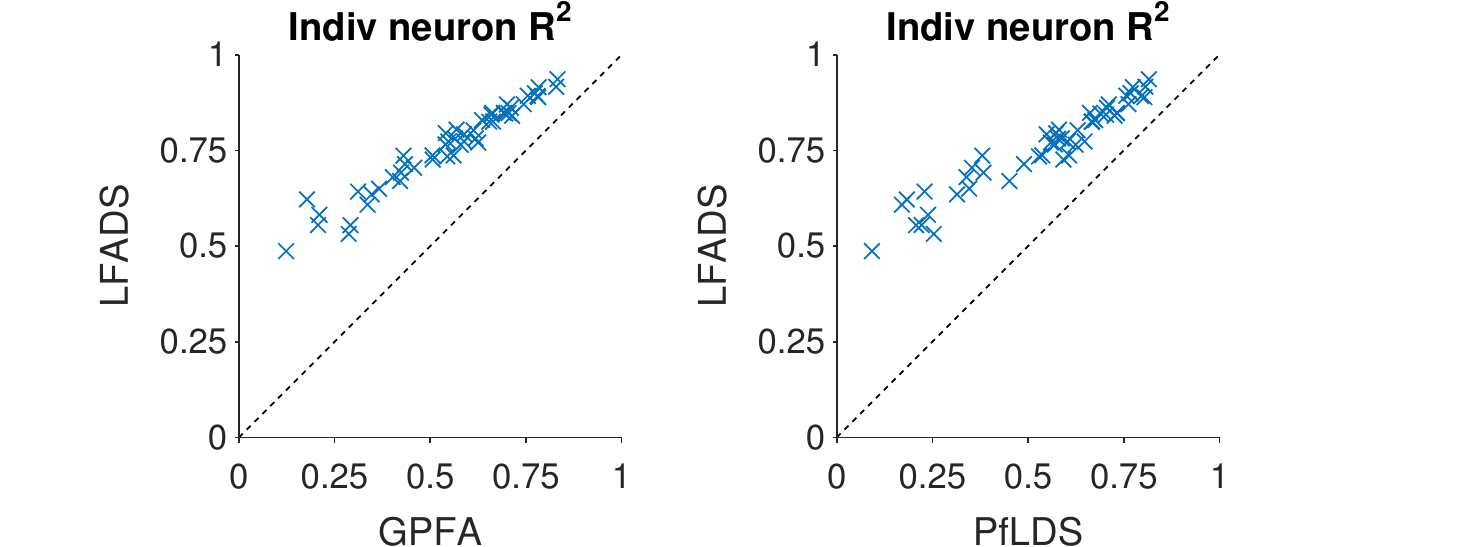}
\caption{ \textbf{$R^2$ values between actual and inferred rates.} Comparison of
  the $R^2$ values for individual neurons from held out data are shown for GPFA
  vs.\ LFADS (left) and PfLDS vs.\ LFADS (right).  In both comparisons, LFADS
  yields a better fit to the data, for every single neuron.}
\label{fig_rnn_r2_individual}
\end{figure*}

We tested the performance of the methods at extracting the underlying firing
rates from the spike trains of the RNN dataset. We restricted this comparison to
the three best-performing methods from the Lorenz comparison (GPFA, PfLDS and
LFADS) because the vLGP results were noticeably worse in that case.  An example
of actual and inferred rates extracted by each model is shown in
Fig.~\ref{fig_rnn_no_thit_wiggles}.  As can be seen by eye, the LFADS results
are closer to the actual underlying rates than for the other models.  We
summarize this again using $R^2$ values between the actual and inferred rates in
Fig.~\ref{fig_rnn_r2_individual}.

\subsection{Inferring inputs to a chaotic RNN}

Finally, we tested the ability of LFADS to infer the input to a dynamical
system, specifically the data RNN used in the previous section. In general, the
problem of disambiguating dynamics from inputs is ill-posed, so we encouraged
the dynamics to be as simple as possible by including an L2 regularizer in the
LFADS network generator (see Appendix).  We note that regularization is a
standard technique that is nearly universally applied to neural network
architectures.  As shown below, adding a regularizer does not mean that the
dynamics play no role.  Rather it ensures that if a low-dimensional inferred
input can aid in explaining the data, the optimization cost will be reduced by
using it instead of building a more complex dynamics.  By low-dimensional
inferred inputs, we mean that the dimensionality of $\uu_t$ should be lower
dimensional than the data we are attempting to model.

\begin{figure*}
\centering
\includegraphics[width=14cm,trim={0 0 3.5cm 0},clip]{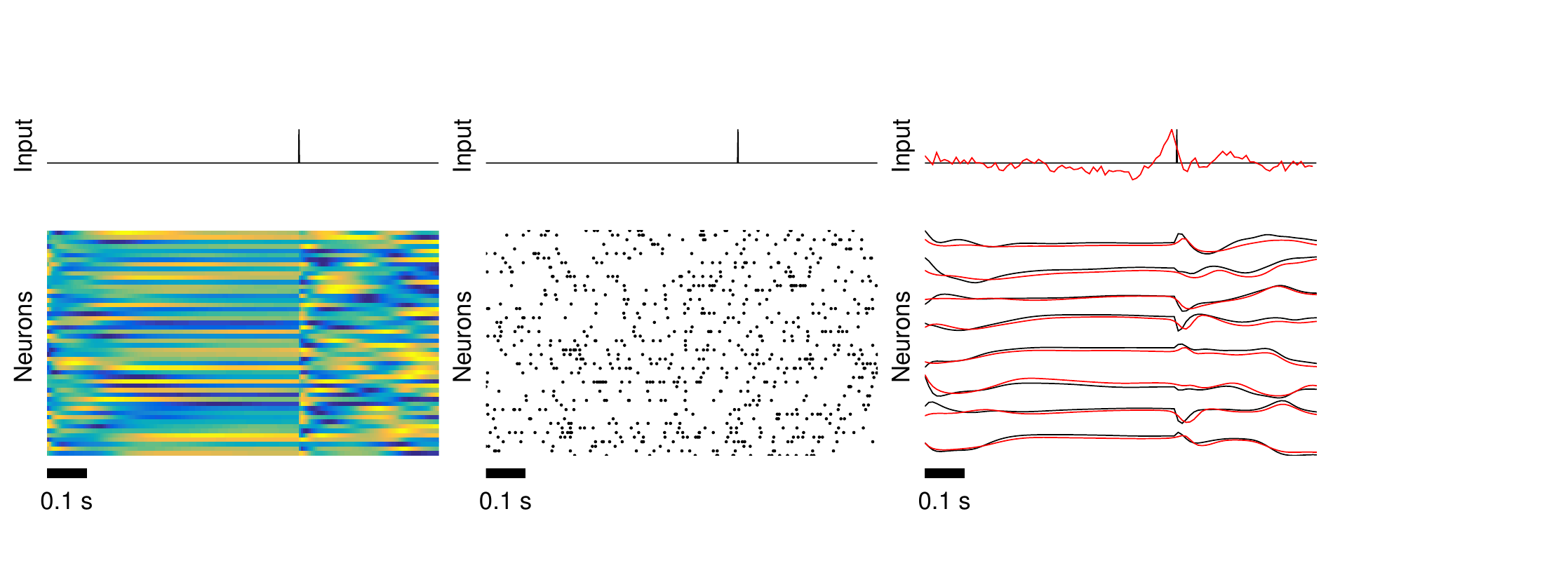}
\caption{ \textbf{Inferring inputs from a data RNN ($\gamma=1.5$) with delta
    pulse inputs.}  Example trial illustrating results from the $\gamma=1.5$
  chaotic data RNN with an external input (shown in black at the top of each
  column).  (Left) Firing rates for the 50 simulated neurons. (Middle)
  Poisson-generated spike times for the simulated neurons. (Right) Example trial
  showing the actual (black) and inferred (red) input (top), and actual firing
  rates of a subset of neurons in black and the corresponding inferred firing
  rates in red (bottom). }
\label{fig_rnn_thit_simple_overview}
\end{figure*}

\begin{figure*}
\centering
\includegraphics[width=14cm,trim={0 0 3.5cm 0},clip]{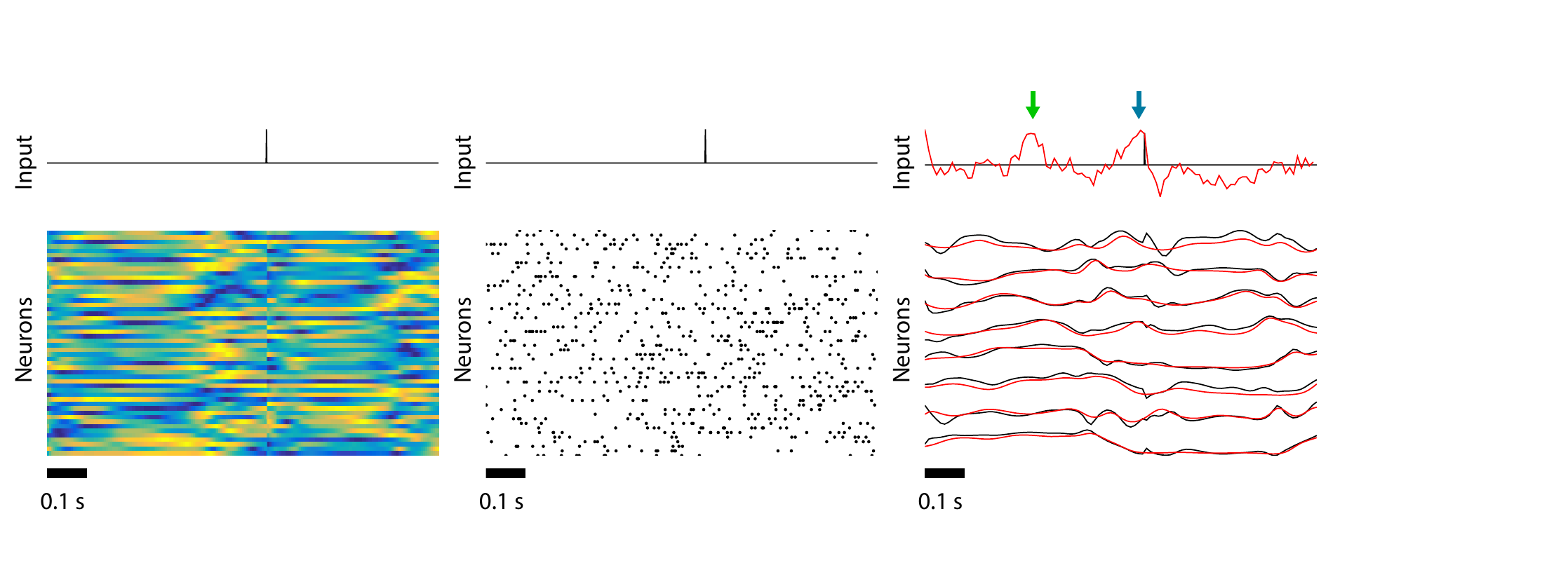}
\caption{ \textbf{Inferring inputs from a data RNN ($\gamma=2.5$) with delta
    pulse inputs.} - Example trial showing results for a driven data RNN with
  more complex dynamics than in Fig.~\ref{fig_rnn_thit_simple_overview}.
  Otherwise, this figure has the same format as
  Fig.~\ref{fig_rnn_thit_simple_overview}. For this more difficult case, LFADS
  inferred the correct input (blue arrow), but also used the input to shape the
  dynamics at times there was no actual input (e.g.\ green arrow).}
\label{fig_rnn_thit_complex_overview}
\end{figure*}

To introduce an input into the data RNN, the elements of $\BB$ were drawn
independently from a normal distribution with zero mean and unit
variance. During each trial, we perturbed the network by delivering a delta
pulse of magnitude 50, $q(t) = 50\delta(t - t_{\mbox{\scriptsize {pulse}}})$, at
a random time $t_{\mbox{\scriptsize {pulse}}}$ between 0.25s and 0.75s (the full
trial length was 1s).  This pulse affects the underlying rates produced by the
data RNN, which modulates the spike generation process.  To test the ability of
the LFADS model to infer the timing of these input pulses, we included in the
LFADS model an inferred input with dimensionality of 1.  We explored two values
of $\gamma$, 1.5 and 2.5\@.  The lower $\gamma$ value produces ``gentler"
chaotic activity in the data RNN than the higher value. Otherwise, the data for
this example (Fig.~\ref{fig_rnn_thit_simple_overview} and
Fig.~\ref{fig_rnn_thit_complex_overview}) were generated as in the first data
RNN example described above.

\begin{figure*}
\centering
\includegraphics[width=14cm]{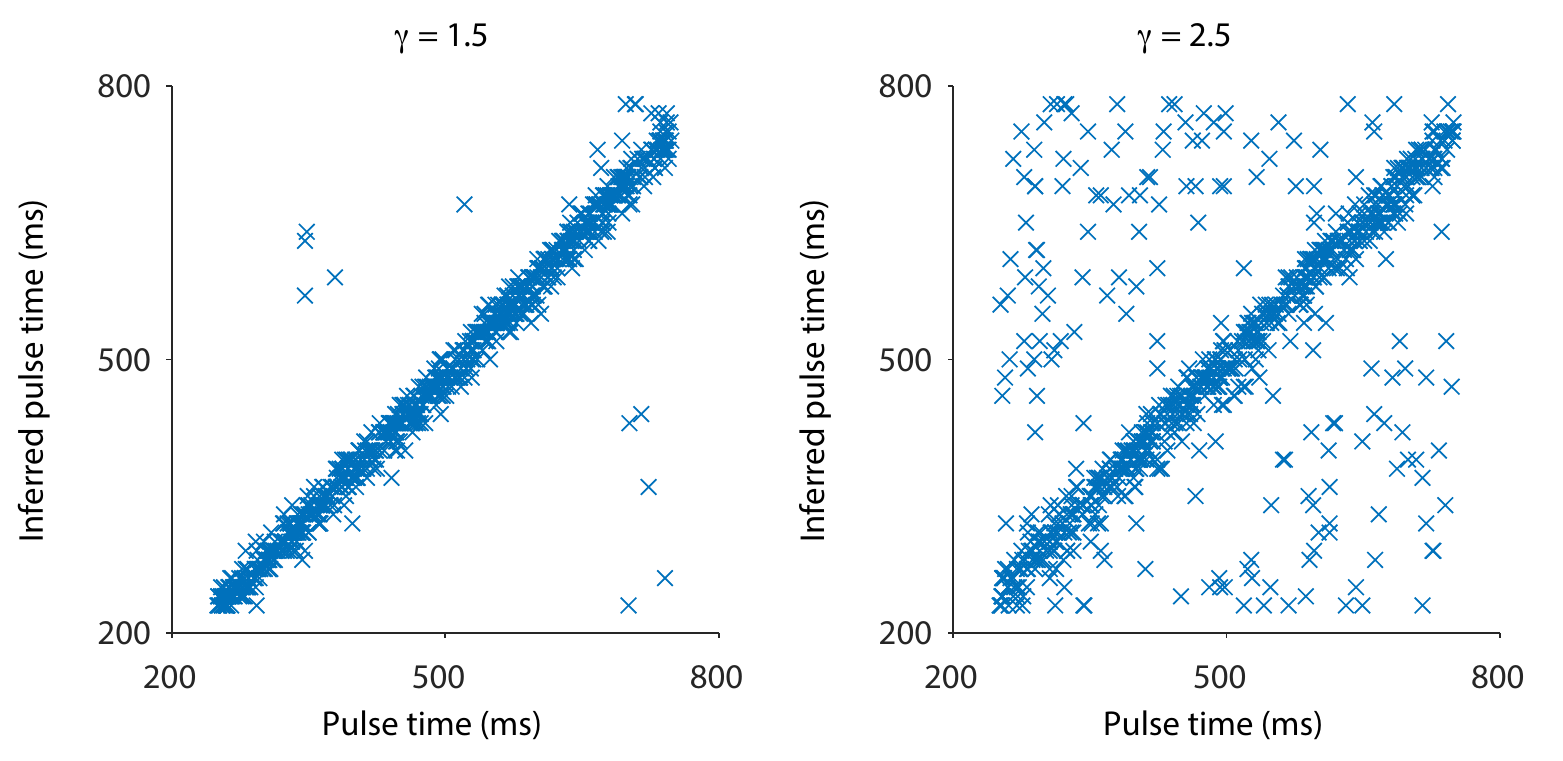}
\caption{ \textbf{Summary of results for inferred inputs.} The inferred time of the input (time of the maximum of $\overline{\uu}_t$; vertical axis) plotted against the actual time of the delta pulse
  (horizontal axis) for all trials.  (Left) $\gamma=1.5$. (Right) $\gamma=2.5$.  These plots show that for the majority of trials, despite complex internal dynamics, LFADS was able to infer the correct timing of a strong input.}
\label{fig_rnn_thit_timing}
\end{figure*}

After training, which successfully inferred the firing rates
(e.g.\ Figs.~\ref{fig_rnn_thit_simple_overview} and~\ref{fig_rnn_thit_complex_overview}, right panels), we extracted
inferred inputs from the LFADS model (eqn. \ref{u_t_sample}) by running the
system 512 times for each trial, and averaging, defining $\overline{\uu}_t =
\langle\uu_t\rangle_{\ggg_0, \uu_{1:T}}$.  To see how this was related to the
actual input pulse, we determined the time at which $\overline{\uu}_t$ reached
its maximum value.  The results are plotted in Fig.~\ref{fig_rnn_thit_timing}
and demonstrate that, for the vast majority of trials, LFADS inferred that there
was an input near the time of the actual delta pulse.

LFADS did a better job of inferring the inputs in the case of simpler dynamics
(i.e., $\gamma=1.5$, Fig.~\ref{fig_rnn_thit_simple_overview}) than for more
complex dynamics ($\gamma=2.5$, Fig.~\ref{fig_rnn_thit_complex_overview}).  This
occurs for two reasons.  First, in the case of $\gamma=2.5$, the complex
dynamics reduces the magnitude of the perturbation caused by the input.  Second,
in the $\gamma=2.5$ case, LFADS used the inferred input more actively to account
for the non-input-driven dynamics as well as the input driven dynamics. The
example of a vanilla RNN driven in the highly chaotic regime ($\gamma = 2.5$)
highlights the subtlety of interpreting an inferred input (see Discussion for
further elaboration on this point).

\begin{figure*}
\centering
\includegraphics[width=14cm]{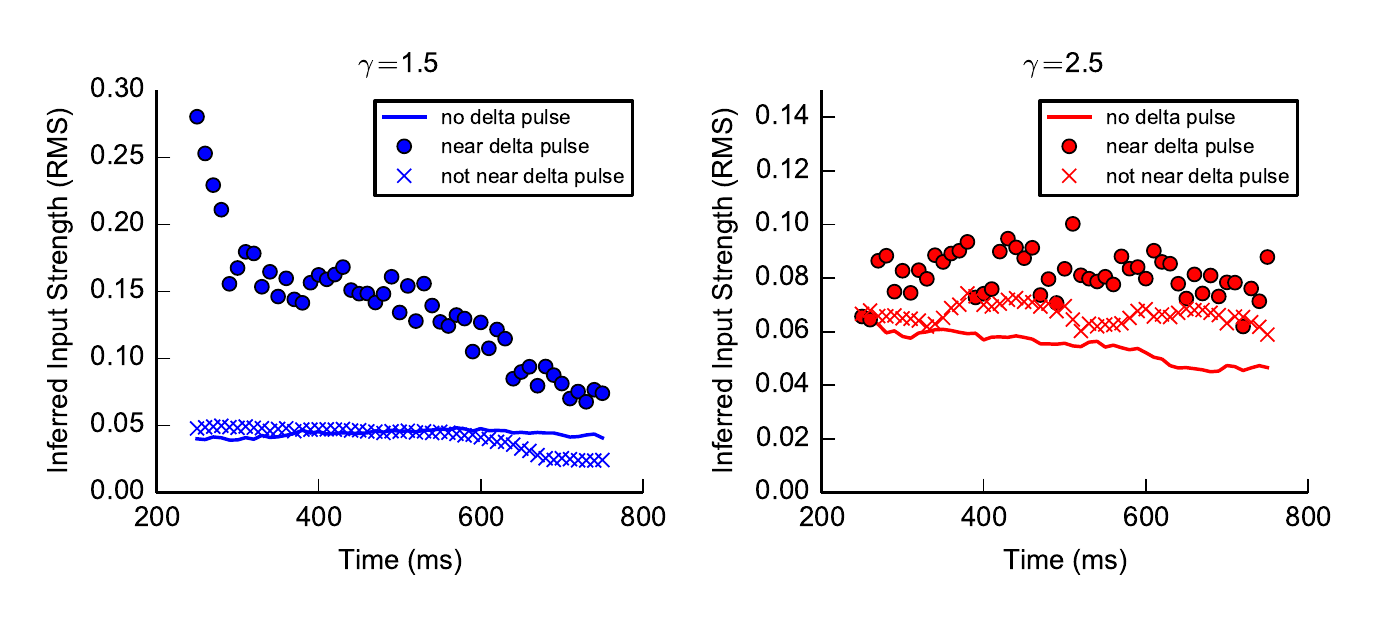}
\caption{ \textbf{Inferred input strength for data RNNs run autonomously
    compared to data RNNs that received delta pulses.} Four data RNNs were
  created, two with $\gamma=1.5$ (left panel, blue), and two with $\gamma=2.5$
  (right panel, red).  For each $\gamma$ value, the data RNN either had no input
  at all or a delta pulse input on each trial.  For each data RNN, we trained an
  LFADS model with a 1-dim inferred input ($u_t$ is a scalar in this
  example). The solid lines show the strength (root-mean-square) of $u_t$ at
  each time point, for the data RNN that received no delta pulses, averaged
  across all examples. The '$\circ$' and 'x' show the strength of $u_t$ for the
  data RNN that received delta pulses, averaged in a time window around $t$,
  and averaged over all examples. Intuitively, a '$\circ$' is the strength of
  $u_t$ around a delta pulse at time $t$, and an 'x' is the strength of $u_t$
  if there was no delta pulse around time $t$. As a reference point, the
  standard deviation of the Gaussian prior for $u_t$ was 0.32.}
  %More precisely, an '$\circ$' represents
  %the strength of $u_t$ in a time window around $t$ when a delta pulse occured
  %at time $t$.  The 'x's represent the strength of $u_t$ in a time window around
  %$t$, if $t$ was outside of the time window used to determine $u_t$ strength at
  %time $t$ when a delta pulse occurred. 
\label{fig_rnn_thit_vs_no_thit}
\end{figure*}

One possibility in using LFADS with inferred inputs (i.e. dimensionality of
$\uu_t \geq 1$) is that the data to be modeled is actually generated by an
autonomous system, yet one, not knowing this fact, allows for an inferred input
in LFADS. To study this case we utilized the four data RNNs described above,
i.e. $\gamma=1.5$, and $\gamma=2.5$, with and without delta pulse inputs.  We
then trained an LFADS model for each of the four cases, with an inferred input
of dimensionality 1, despite the fact that two of the four data RNNs generated
their data autonomously.  We trained all four models, and after training we
examined the strength of the average inferred input, for each LFADS model
($\overline{\uu}_t$ as defined above).  Our definition of strength is
root-mean-square of the inferred input, averaged over an appropriate time
window, $\sqrt{\langle\overline{\uu}_t^2\rangle_t}$.  The results and details of
this analysis are shown in Fig.~\ref{fig_rnn_thit_vs_no_thit}.  Importantly, the
strength of the inferred input when pulses were not present in the data was
similar to the magnitude of inferred input when pulses were present in the data
but not in the specific window.  Further, when inputs were present in the data
and within the specific window, the magnitude of the inferred input was higher
on average than cases without inputs.


\section{Discussion}

We have developed the LFADS algorithm for inferring latent dynamics from
single-trial, high-dimensional neural spike trains.  We tested the algorithm on
synthetic datasets based on the nonlinear Lorenz attractor and a chaotic
nonlinear RNN and showed that LFADS significantly outperforms other
state-of-the-art algorithms in these examples.  We note that if one wishes to
study linear systems, a variety of optimal methods exist in the literature
(e.g. Kalman filter).

The only explicit aspect of LFADS that ties it to neuroscience is the Poisson
spiking process.  By exchanging the Poisson distribution for a Gaussian emission
distribution, for example, LFADS could be applied to all manner of time series
data.  However, there are further assumptions that implicitly tailor LFADS to
neural data, which we outline in the next few paragraphs.  For example, we use
the temporal factors to implement a dimensionality bottleneck between the
generator and the rates.  This is based on the commonly observed phenomenon in
neuroscience that the number of recorded units typically far exceeds the
dimensionality of the data \cite{gao2015simplicity}.

In many machine learning applications, all the inputs to the system are known.
A particular aspect of neuroscience data is that the inputs typically are not
known, so they must be inferred.  (For example, even in primary visual cortex,
the visual input from the world has already been transformed a number of times.)
From a theoretical perspective, it may be extremely difficult to correctly
factor observed data into features generated by a dynamical system and features
generated by an feed-forward time-varying input, because the initial conditons
or the feed-forward input are incompletely known.  In applying LFADS to
neuroscience data, regularizers on the complexity of either the dynamical system
or the inferred inputs could be used to determine the relative weighting of
these two factors.  For example, in motor tasks network dynamics may be
predominant, while inferred input may be more important in studies with
extensive sensory input.  In a neuroscience setting where input is so difficult
to measure, it is not unreasonable to attempt to infer inputs, even if the
results are only qualitatively correct, but caution should be exercised.

Often in deep sequence models, such as language models, the emission
distribution (typically a softmax over discrete symbols) is viewed as a
fundamental source of structural variability in the sequence.  As a result, when
these models are used for inference, a sample of the emission process is fed
back to the network.  In the case of LFADS applied to spiking neural data, we
view the spikes as noisy observations of an underlying smooth process, so
instantiating spikes and feeding them back into the generator would only serve
to inject noise into the system (by assumption), not enhance structural
variability. Thus, we do not do it.  If one wanted to adapt LFADS to an
application where the emission was not viewed as noise, then one would sample
$\hat{\xx}_{t} \sim\ P(\xx_{t}|...)$ and feed that to the generator and
controller at the next time step, instead of $\ff_{t}$ injected only to the
controller.

While the primary motivation for the LFADS algorithm was inferring smooth latent
variables, there is no reason why LFADS should be restricted to slow dynamics.
The discrete-time RNN used in the algorithm can generate fast dynamics, so the
fastest timescale that LFADS can produce is only limited by the binning or
temporal resolution of the experimental spike counts. Relevant to this point, a
clear sign that LFADS is beginning to overfit the data is the production of
``spiky'' rates with high-frequency changes that do not generalize to the
held-out data, especially near the beginning of a trial or immediately after a
strong input.

There are some obvious extensions and future directions to explore. First, an
emissions model relevant to calcium imaging would be extremely useful for
inferring neural firing rate dynamics underlying calcium signals.  Second, in
this contribution we implemented LFADS as a ``smoother'', in Kalman filter
language, that cannot run in real time.  It would be interesting to adapt LFADS
as a ``filter'' that could.  Third, the LFADS generator could clearly be
strengthed by stacking recurrent layers or adding a feed-forward deep net before
the emissions distribution at each time step.  Another extension would be to
learn the dimensions of the inferred input and temporal factors automatically,
instead of having them specified as predetermined hyper-parameters (e.g. nuclear
norm minimization on the respective matrices). Finally, we can imagine an
LFADS-type algorithm that leans more towards the feed-forward side of
computation, but still has some recurrence.  An application would be, for
example, to explain short-term effects in visual processing.  In this setting,
the information transfer along the temporal dimension could be limited while
expanding the information flow in the feed-forward direction.
% \TODO{Is this the Pfau citation}

\section{Acknowledgments}

We would like to thank John P. Cunningham, Laurent Dinh, and Jascha
Sohl-Dickstein for extensive conversation.  R.J. participated in this work while
at Google, Inc.  L.F.A.’s research was supported by US National Institutes of
Health grant MH093338, the Gatsby Charitable Foundation through the Gatsby
Initiative in Brain Circuitry at Columbia University, the Simons Foundation, the
Swartz Foundation, the Harold and Leila Y. Mathers Foundation, and the Kavli
Institute for Brain Science at Columbia University.  C.P. was supported by a
postdoctoral fellowship from the Craig H. Neilsen Foundation for spinal cord
injury research.

\section{Appendix}

%% \subsection{Reference implementation}
%% An open source reference implementation of LFADS will be made available at \newline
%% \url{https://github.com/tensorflow/models/lfads}.}

\subsection{GRU equations}
The GRU equations are as follows,
\begin{align}
  \rr_t &= \sigma\left(\WW^{r}(\left[\xx_t, \hh_{t-1}\right])\right) \label{gru_eq_1} \\ 
  \uu_t &= \sigma\left(\WW^{u}(\left[\xx_t, \hh_{t-1}\right])\right) \\
  \cc_t &= \tanh\left(\WW^{c}(\left[\xx_t, \rr_t \odot \hh_{t-1}\right])\right) \\
  \hh_t &= \uu_t \odot \hh_{t-1} + (1-\uu_t) \odot \cc_t, \label{gru_eq_4}
\end{align}
with $\hh_t$ being the hidden state, $\odot$ denoting element-wise
multiplication and $\sigma$ denoting the logistic function. For clarity, we use
the common variable symbols associated with the GRU, with the understanding
that the variables represented here by these symbols are not the same variables as
those in the general LFADS model description.

\subsection{Further details of LFADS implementation.}
While the general model is defined in the main manuscript, there were a number of
details that aided in the optimization and generalization of the LFADS model
applied to the datasets.  
\begin{itemize}
\item The matrix in the $\WW^{\fac}(\cdot)$ affine transformation is
  row-normalized to keep the factors relatively evenly scaled with respect to
  each other.
\item Following the authors in \cite{bowman2016generating}, we added a schedule
  on the KL divergence penalty so that the optimization does not quickly set the KL
  divergence to 0\@.
\item We experimented with the variance of the prior distribution for both the
  initial condition and inferred input distributions and settled on a value of
  0.1, chosen to avoid saturating network nonlinearities.
\item To help avoid over-fitting, we added a dropout layer
  \cite{hinton2012improving} to the inputs and to a few feed-forward
  (input) connections \cite{zaremba2014recurrent} in the LFADS
  model. Specifically, we used dropout ``layers'' around
  equation~\ref{eqn_enc_3}, around the input in
  equation~\ref{eqn_dropout_2}, and around
  equation~\ref{eqn_dropout_3}.
\item We added an $L_2$ penalty to recurrent portions of the generator
  (equations~\ref{gru_eq_1}-\ref{gru_eq_4}) to encourage simple
  dynamics. Specifically, we regularized any matrix parameter by which
  $\hh_{t-1}$ was multiplied, but not those that multiplied $\xx_t$.
\item As defined in eqn. \ref{eq_L_KL}, there is an information limiting regularizer
  placed on $\uu_t$ by virtue of minimizing the KL divergence between
  the approximate posterior over $\uu_t$ and the uninformative Gaussian prior.
\item We clipped our hidden state $\hh_t$ when any of its values went above a set
  threshold and used gradient clipping to avoid occasional pathological
  gradients.
\end{itemize}
\small
\bibliographystyle{acm}
\bibliography{lfads}

%\end{multicols}
\end{document}